\documentclass[runningheads,a4paper]{llncs}

\usepackage{amssymb}
\usepackage{amsmath}
\usepackage{url}
\usepackage{graphicx}
\usepackage{multirow}
\usepackage{wrapfig}
\usepackage{hyperref}
\usepackage{color}
\usepackage{colortbl}
\usepackage[font=small,labelfont=bf]{caption}
\usepackage[table]{xcolor}

\definecolor{mygray}{gray}{.9}
\makeatletter
\newcommand{\thickhline}{%
    \noalign {\ifnum 0=`}\fi \hrule height 1pt
    \futurelet \reserved@a \@xhline
}
\makeatother

\newcommand{\ie}{{\it i.e.}}

\mathchardef\mhyphen="2D
\begin{document}

\mainmatter  % start of an individual contribution

\title{ET-Net: A Generic Edge-aTtention Guidance Network for Medical Image Segmentation}

% \titlerunning{ET-Net}


% anonymous stuff
\author{Zhijie Zhang \inst{1}, Huazhu Fu \inst{2}\thanks{Corresponding author (\url{huazhu.fu@inceptioniai.org}).}, Hang Dai \inst{2}, Jianbing Shen \inst{2}, \\ Yanwei Pang \inst{1} and Ling Shao \inst{2}}

% index{Zhang, Zhijie}
% index{Fu, Huazhu}
% index{Dai, Hang}
% index{Shen, Jianbing}
% index{Pang, Yanwei}
% index{Shao, Ling}

\authorrunning{Z Zhang, H Fu, H Dai, J Shen, Y Pang and L Shao}
\tocauthor{Zhijie Zhang, Huazhu Fu, Hang Dai, Jianbing Shen, Yanwei Pang and Ling Shao}
\institute{School of Electrical and Information Engineering, Tianjin University \and Inception Institute of Artificial Intelligence, UAE}

\maketitle

%%%%%%
\begin{abstract}
Segmentation is a fundamental task in medical image analysis. However, most existing methods focus on primary region extraction and ignore edge information, which is useful for obtaining accurate segmentation. In this paper, we propose a generic medical segmentation method, called Edge-aTtention guidance Network (ET-Net), which embeds edge-attention representations to guide the segmentation network. Specifically, an edge guidance module is utilized to learn the edge-attention representations in the early encoding layers, which are then transferred to the multi-scale decoding layers, fused using a weighted aggregation module. The experimental results on four segmentation tasks (\ie, optic disc/cup and vessel segmentation in retinal images, and lung segmentation in chest X-Ray and CT images)  demonstrate that preserving edge-attention representations contributes to the final segmentation accuracy, and our proposed method outperforms current state-of-the-art segmentation methods. The source code of our method is available at~\url{https://github.com/ZzzJzzZ/ETNet}.

\end{abstract}
%%%%%%

\section{Introduction}
\label{sec:introdution}
Medical image segmentation is an important procedure in medical image analysis. The shapes, size measurements and total areas of segmentation outcomes can provide significant insight into early manifestations of life-threatening diseases. As a result, designing an efficient general segmentation model deserves further attention. Existing medical image segmentation methods can mainly be divided into two categories: edge detection and object segmentation. The edge detection methods first identify object boundaries utilizing local gradient representations, and then separate the closed loop regions as the objects. These methods, which aim to obtain highly localized image information, can achieve high accuracy in boundary segmentation, and are adequate for simple structures. For example, the level-set technique is employed to minimize an objective function for estimating tumor segmentation based on shape priors~\cite{levelset2003}. The template matching method is proposed to obtain optic disc boundary approximations with the Circular Hough Transform in retinal images~\cite{Aquino2010}. Other edge detection methods are employed to extract blood vessels in retinal images~\cite{deepvessel2016,MOCCIA201871}. However, these edge detection methods depend on local edge representations and lack object-level information, which leads to trivial segmentation regions and discontinuous boundaries. By contrast, object segmentation methods~\cite{salient,Gu2019} utilize global appearance models of foregrounds and backgrounds to identify the target regions, which preserves the homogeneity and semantic characteristics of objects, and reduces the uncertainties in detecting the boundary positions. A common way of doing this is to classify each pixel/patch in an image as foreground or background. For example, a superpixel classification method was proposed to segment the optic disc and cup regions for glaucoma screening~\cite{Cheng2013}.  However, without utilizing edge information, several object segmentation methods need to refine the initial coarse segmentation results using additional post-processing technologies (\textit{e.g.}, Conditional Random Field and shape fitting), which is time-consuming and less related to previous segmentation representations.

Recently, the success of U-Net has significantly promoted widespread applications of segmentation on medical images, such as cell detection from 2D image~\cite{Ronneberger2015}, vessel segmentation from retinal images~\cite{deepvessel2016} and lung region extraction from chest X-Ray and CT images~\cite{lung2015}. However, there are still several limitations when applying Deep Convolutional Neural Networks (DCNNs) based on a U-Net structure. In medical image segmentation, different targets sometimes have similar appearances, making it difficult to segment them using a U-Net based DCNN. Besides, inconspicuous objects are sometimes over-shadowed by irrelevant salient objects, which can confuse the DCNNs, since it cannot extract discriminative context features, leading to false predictions. In addition, target shapes and scale variations are difficult for DCNNs to predict. Although U-Net proposes to aggregate high-level and low-level features to address this problem, it only slightly alleviates it, since it aggregates features of different scale without considering their different contributes. Herein, we propose a novel method to extract discriminative context features and selectively aggregate multi-scale information for efficient medical image segmentation.

In this paper, we integrate both edge detection and object segmentation in one deep learning network. To do so, we propose a novel general medical segmentation method, called Edge-aTtention guidance Network (ET-Net), which embeds edge-attention representations to guide the process of segmentation.
In our ET-Net, an edge guidance module (EGM) is provided to learn the edge-attention representations and preserve the local edge characteristics in the early encoding layers, while a weighted aggregation module (WAM) is designed to aggregate the multi-scale side-outputs from the decoding layers and transfer the edge-attention representations to high-level layers to improve the final results. We evaluate the proposed method on four segmentation tasks, including optic disc/cup segmentation and vessel detection in retinal images, and lung segmentation in Chest X-Ray and CT images. Results demonstrate that the proposed ET-Net outperforms the state-of-the-art methods in all tasks.

\section{Method}
\label{sec:method}

\begin{figure}[!t]
	\centering
	\includegraphics[width=1.0\linewidth]{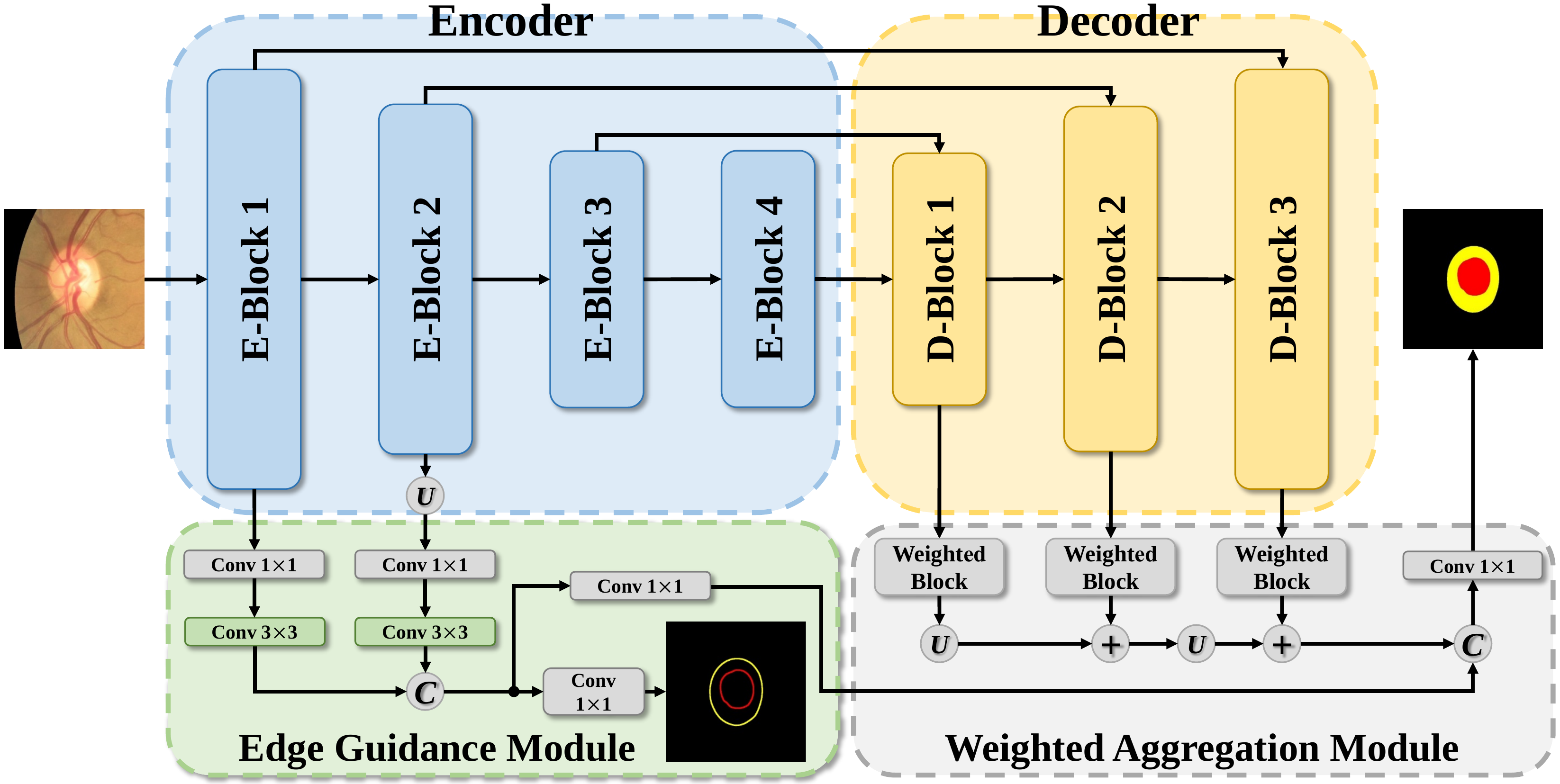}
	\caption{Illustration of our ET-Net architecture, which includes the main encoder-decoder network with an edge guidance module and weighted aggregation module. `Conv'  denotes the convolutional layer, while `U', `C', and `+' denote the upsampling, concatenation, and addition layers, respectively.}
	\label{fig_network}
\end{figure}

Fig.~\ref{fig_network} illustrates the architecture of our ET-Net, which is primarily based on an encoder-decoder network, with the EGM and WAM modules appended on the end. The ResNet-50~\cite{resnet2016he} is utilized as the encoder network, which comprises of four Encoding-Blocks (E-Blocks), one for each different feature map resolution. For each E-Block, the inputs first go through a feature extraction stream, which consists of a stack of $1\!\times\!1 - 3\!\times\!3 - 1\!\times\!1$ convolutional layers, and are then summed with the shortcut of inputs to generate the final outputs. With this residual connection, the model can generate class-specific high-level features. The decoder path is formed from three cascaded Decoding-Blocks (D-Blocks), which are used to maintain the characteristics of the high-level features from the E-Blocks and enhance their representation ability. As shown in Fig.~\ref{fig:module} (a), the D-Block first adopts a depth-wise convolution to enhance the representation of the fused low-level and high-level features.  Then, a $1\!\times\!1$ convolution is used to unify the number of channels.

\subsection{Edge Guidance Module}
\label{sec:method:EGM}

As stated in Sec.~\ref{sec:introdution}, edge information provides useful fine-grained constraints to guide feature extraction during segmentation. However, only low-level features preserve sufficient edge information. As such, we only apply the EGM at the top of early layers, \ie, E-Block 1 and 2, of the decoding path, as shown in Fig.~\ref{fig_network}.  The EGM has two main fashions: 1) it provides an edge-attention representation to guide the process of segmentation in the decoding path; 2) it supervises the early convolutional layers using the edge detection loss.

In our EGM, the outputs of E-Block 2 are upsampled to the same resolution as the outputs of E-Block 1, and then fed into the $1\!\times\!1 - 3\!\times\!3$ convolutional layers and concatenated together. After that, the concatenated features go through one of two branches: a $1\!\times\!1$ convolutional layer to act as the edge guidance features in the decoding path, or another $1\!\times\!1$ convolutional layer to predict the edge detection results for early supervision.
The Lov\'{a}sz-Softmax loss~\cite{Berman_2018_CVPR} is used in our EGM, since it performs better than cross entropy loss for class imbalanced problems. It can be formulated as:
\begin{equation}
\mathcal{L}  = \frac{1}{C}  \sum\limits_{c \in C} \overline{\Delta_{J_c}} (m(c)),
\;\;\; \text{and} \;\;\;
    m_i(c) = \left\{
        \begin{array}{ll}
          1-p_i(c) \;\; & \text{if} \;\; c=y_i(c),  \\
          p_i(c)   & \text{otherwise},
        \end{array}
    \right.
\end{equation}
where $C$ denotes the class number, and $y_i(c) \in \{-1, 1\}$ and $p_i(c) \in [0, 1]$ are the ground truth label and predicted probability of pixel $i$  for class $c$, respectively. $\overline{\Delta_{J_c}}$ is the Lov\'{a}sz extension of the Jaccard index~\cite{Berman_2018_CVPR}. With the edge supervision, the transmitted edge features are better able to guide the extraction of discriminative features in high-level layers.

\subsection{Weighted Aggregation Module}
\label{sec:method:WAM}

\begin{figure}[!t]
	\centering
	\includegraphics[width=1.0\linewidth]{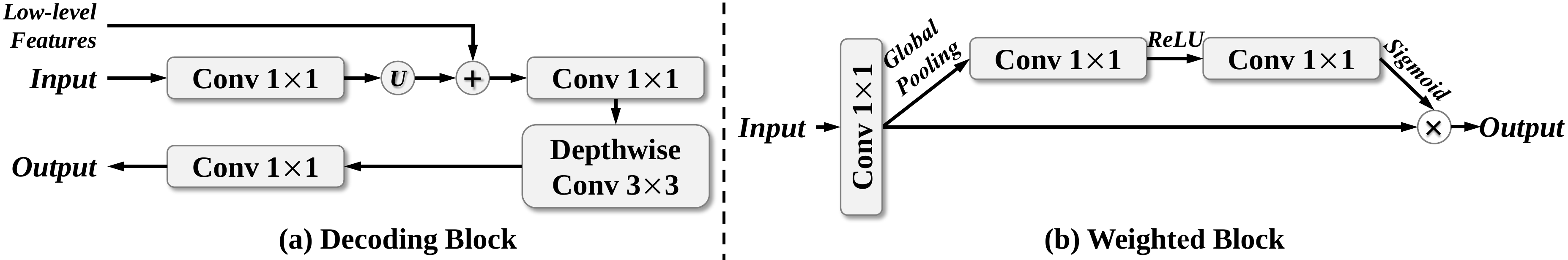}
	\caption{Illustration of the E-Block and Weighted Block. `U', `+' and `$\times$' denote upsampling, addition, and multiplication layers, respectively.}
	\label{fig:module}
\end{figure}

In order to adapt to the shape and size variations of objects, existing methods tend to sum up multi-scale outputs along the channel dimension for final predictions (\textit{e.g.},~\cite{mnet2018fu,shen2019}). However, not all features in high-level layers are activated and benefit the recovery of objects. Aiming to address this, we develop the WAM to emphasize the valuable features, and aggregate multi-scale information and edge-attention representations to improve the segmentation performance.
As shown in Fig.~\ref{fig_network}, outputs of each D-Block are fed into the Weighted Blocks to highlight the valuable information.
The structure of a Weighted Block is shown in Fig.~\ref{fig:module}(b). In this block, global average pooling is first employed to aggregate the global context information of inputs, and then two $1\!\times\!1$ convolutional layers with different non-linearity activation functions, \ie, ReLU and Sigmoid, are applied to estimate the layer relevance and generate the weights along the channel dimension. After that, the generated weights are multiplied with the outputs to yield more representative features.

Our WAM integrates the features of different scales via a bottom-up pathway, which generates a feature hierarchy consisting of feature maps of different sizes. Finally, the WAM also concatenates edge-attention representations from the EGM, and then applies a $1\!\times\!1$ convolution to extract features under edge-guided conditions. As with the edge detection in the EGM, our WAM also utilizes the Lov\'{a}sz-Softmax loss as the segmentation loss function. Thus, the total loss function of our ET-Net is defined as: $\mathcal{L}_{total}=\alpha \cdot \mathcal{L}_{seg}+ (1-\alpha) \cdot \mathcal{L}_{edge}$,
where $\mathcal{L}_{edge}$ and $\mathcal{L}_{seg}$ denote the losses for edge detection in EGM and segmentation in WAM, respectively. In our experiments, weight $\alpha$ is empirically set to 0.3.

\subsection{Implementation Details}

For data augmentation, we apply a \textit{random mirror}, \textit{random scale}, which ranges from 0.5 to 2, and \textit{random rotation} between -10 and 10 degrees, for all datasets. \textit{random color jitters} with a probability of 0.5 are also applied to the data. All input images are randomly cropped to $512\!\times\!512$.

The initial weights of the encoder network come from ResNet-50~\cite{resnet2016he}  pre-trained on ImageNet, and the parameters of the other layers are randomly initialized.
A dilation strategy is used in E-Block 4, with an output stride of 1/16. During training, we set the \textit{batch\_size} to 16 with synchronized batch normalization, and adopt ‘poly’ learning rate scheduling $lr = base\_lr \times (1-\frac{iters}{total\_iters})^{power}$, in which the power is set to 0.9 and \textit{base\_lr} is 0.005. The \textit{total\_iters} is calculated by the $ num\_images \times epochs / batch\_size$, where \textit{epochs} is set to 300 for all datasets. Our deep models are optimized using the Adam optimizer with a momentum of 0.9 and a weight decay of 0.0005. The whole ET-Net framework is implemented using PyTorch. Training (300 epochs) requires approximately 2.5 hours on one NVIDIA Titan Xp GPU. During testing, the segmentation results, including edge detection and object segmentation, are produced within 0.015 sec. per image.

%%%
\section{Experiments}
\label{sec:exp}

We evaluate our approach on three major types of medical images: retinal images, X-Ray and CT images. For convenience comparison, we select the evaluation metrics that are highly related to generic segmentation.

\begin{table}[!t]
\footnotesize
\begin{center}
\caption{Optic disc/cup segmentation results on retinal fundus images.}
\begin{tabular}{c|c|c|c|c|c|c}
	\hline
	             Method               &         \multicolumn{3}{c|}{\textbf{REFUGE}}         &       \multicolumn{3}{c}{\textbf{Drishti-GS}}        \\ \cline{2-7}
	                                  & Dice$_{OC}$ (\%) & Dice$_{OD}$ (\%) &   mIoU (\%)    & Dice$_{OC}$ (\%) & Dice$_{OD}$ (\%) &   mIoU (\%)    \\ \hline
	         FCN~\cite{fcn}           &      84.67       &      92.56       &     82.47      &      87.95       &      95.69       &     83.92      \\
	  U-Net~\cite{Ronneberger2015}    &      85.44       &      93.08       &     83.12      &      88.06       &      96.43       &     84.87      \\
	     M-Net~\cite{mnet2018fu}      &      86.48       &      93.59       &     84.02      &      88.60       &      96.58       &     85.88      \\
	   Multi-task~\cite{multitask}    &      86.74       &      94.01       &     84.36      &      88.96       &      96.55       &     85.94      \\
	\textit{p}OSAL~\cite{Wang2019tmi} &      87.50       &      94.60       &       -        &      90.10       &      97.40       &       -        \\ \hline
	           Our ET-Net             &  \textbf{89.12}  &  \textbf{95.29}  & \textbf{86.70} &  \textbf{93.14}  &  \textbf{97.52}  & \textbf{87.92} \\ \hline
\end{tabular}
\label{tab:fundus}
\end{center}
\end{table}

\noindent\textbf{Optic disc \& cup segmentation in retinal images:} We evaluate our method on optic disc and cup segmentation in retinal images, which is a common task in glaucoma detection. Two public datasets are used in this experiment: the REFUGE\footnote{\href{https://refuge.grand-challenge.org/}{https://refuge.grand-challenge.org/}} dataset, which consists of 400 training images and 400 validation images; the Drishti-GS~\cite{Drishti} dataset, which contains 50 training images and 51 validation images. Considering the negative influence of non-target areas in fundus images, we first localize the disc centers following the existing automatic disc detection method~\cite{mnet2018fu}, and then transmit the localized images into our network.
The proposed approach is compared with the classic segmentation methods (\ie, FCN~\cite{fcn} U-Net~\cite{Ronneberger2015}, M-Net~\cite{mnet2018fu}, and Multi-task~\cite{multitask} (it predicts edge and object predictions on the same features.), and the state-of-the-art segmentation method \textit{p}OSAL~\cite{Wang2019tmi}, which achieved first place for the optic disc and cup segmentation tasks in the REFUGE challenge. The dice coefficients of optic disc (Dice$_{OD}$) and cup (Dice$_{OD}$), as well as mean intersection-over-union (mIoU), are employed as evaluation metrics.
As shown in Table~\ref{tab:fundus}, our ET-Net achieves the best performance on both the REFUGE and Drishti-GS datasets. Our model achieves particularly impressive results for optic cup segmentation, which is an especially difficult task, achieving 2\% improvement of Dice$_{OC}$ over the next best method.

\begin{table}[!t]
	\footnotesize
	\begin{center}
		\caption{Segmentation results on retinal fundus, X-Ray and CT images.}
		\begin{tabular}{c|c|c|c|c|c|c}
			\hline
			\textbf{Method}  &    \multicolumn{2}{c|}{\textbf{DRIVE}}&    \multicolumn{2}{c|}{\textbf{MC}}    &   \multicolumn{2}{c}{\textbf{LUNA}}    \\ \cline{2-7}
			                & \textbf{ Acc.(\%)} & \textbf{mIoU(\%)} & \textbf{Acc.(\%)} & \textbf{mIoU(\%)}& \textbf{Acc.(\%)} & \textbf{mIoU(\%)} \\ \hline
			     FCN~\cite{fcn}     & 94.13 & 74.55 & 97.35  & 90.53 & 96.18 & 93.82 \\
			     U-Net~\cite{Ronneberger2015}     & 94.45 & 75.46 & 97.82 & 91.64 & 96.63 & 94.79 \\
			     M-Net~\cite{mnet2018fu}  & 94.58 & 75.81 & 97.96 & 91.95 & 97.27 & 94.92 \\
			     Multi-task~\cite{multitask}     & 94.97 & 76.21 & 98.13 & 92.24 & 97.82 & 94.96 \\ \hline
			  Our ET-Net   & \textbf{95.60} & \textbf{77.44} &  \textbf{98.65} & \textbf{94.20} & \textbf{98.68}   & \textbf{96.23} \\ \hline
		\end{tabular}
		\label{tab:lung}
	\end{center}
\end{table}

\noindent\textbf{Vessel segmentation in retinal images:} We evaluate our method on vessel segmentation in retinal images. DRIVE~\cite{DRIVE}, which contains 20 images for training and 20 for testing, is adopted in our experiments. The statistics in Table~\ref{tab:lung} show that our proposed method achieves the best performance, with 77.44\% mIoU and 95.60\% accuracy, when compared with classical methods (\ie, U-Net~\cite{Ronneberger2015}, M-Net~\cite{mnet2018fu}, FCN~\cite{fcn} and Multi-task~\cite{multitask}).

\noindent\textbf{Lung segmentation in X-Ray images:} We conduct lung segmentation experiments on Chest X-Rays, which is an important component for computer-aided diagnosis of lung health. We use the Montgomery County (MC)~\cite{mc} dataset, which contains 80 training images and 58 testing images. We compare our ET-Net with FCN~\cite{fcn}, U-Net~\cite{Ronneberger2015}, M-Net~\cite{mnet2018fu} and Multi-task~\cite{multitask}, in terms of mIoU and accuracy (Acc.) scores. Table~\ref{tab:lung} shows the results, where our method achieves the state-of-the-art performance, with an Acc. of 98.65\% and mIoU of 94.20\%.

\noindent\textbf{Lung segmentation in CT images:} We evaluate our method on lung segmentation from CT images, which is fundamental for further lung nodule disease diagnosis. The Lung Nodule Analysis (LUNA) competition dataset\footnote{\href{https://www.kaggle.com/kmader/finding-lungs-in-ct-data/data}{https://www.kaggle.com/kmader/finding-lungs-in-ct-data/data}} is employed, which is divided into 214 images for training and 53 images for testing. As with the lung segmentation from Chest X-Ray images, we compare our method with FCN~\cite{fcn}, U-Net~\cite{Ronneberger2015}, M-Net~\cite{mnet2018fu}, and Multi-task~\cite{multitask}, in terms of mIoU and Acc. scores. The randomly cropped images are fed into the proposed network. As shown in Table~\ref{tab:lung}, our ET-Net outperforms previous state-of-the-art methods, obtaining an Acc. of 98.68\% and mIoU of 96.23\%.

%It is noted that, all the experimental results are evaluated directly from the network predictions without any post processing, which means the proposed method is really end-to-end network with the state-of-the-art performance.

\begin{figure}[!t]
\centering
\includegraphics[width=1.0\linewidth]{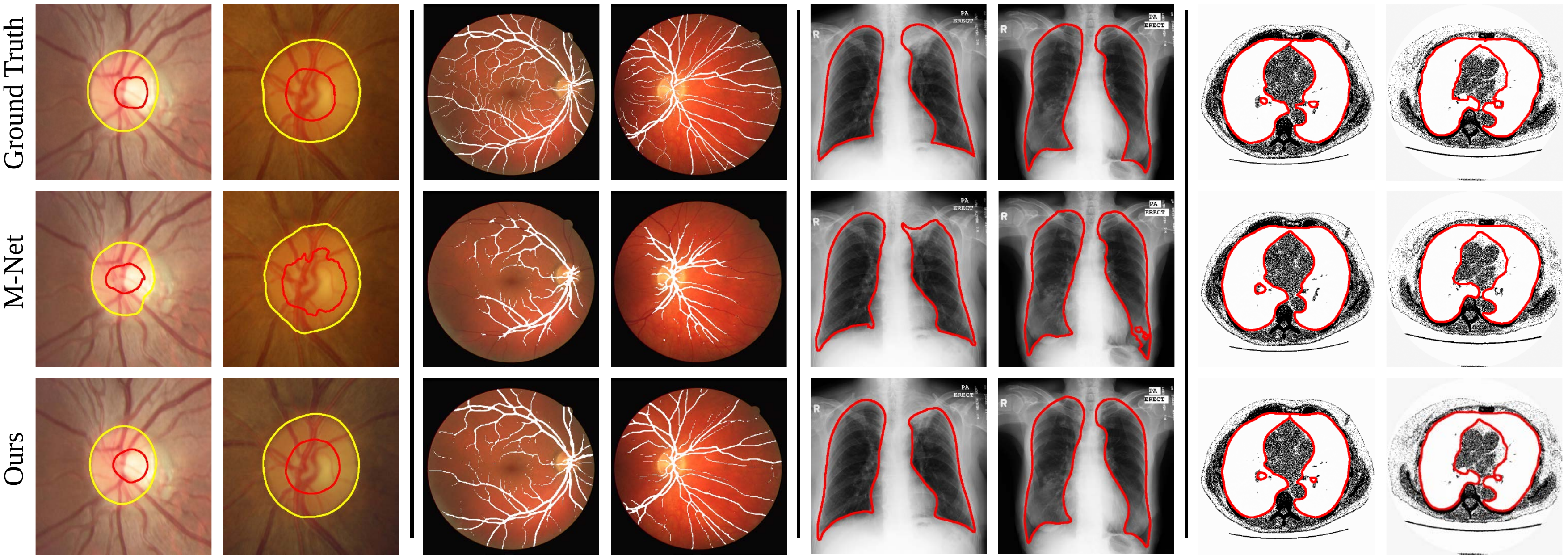}
\caption{Visualization of segmentation results. From left to right: optic disc/cup, and vessel segmentation in retinal fundus images, lung segmentation in Chest X-Ray and CT images.}
\label{fig:reports}
\end{figure}

In addition to quantitative results, we provide qualitative segmentation results, shown in Fig.~\ref{fig:reports}. As can be seen, our results are close to the ground truth. When compared with the predictions of other methods, it is clear that our results are better, especially, in the edge regions.

\begin{table}[!t]
\footnotesize
\begin{center}
\caption{Ablation study of optic disc/cup segmentation on the Drishti-GS \textit{test} set}
\setlength{\tabcolsep}{3pt}{
\begin{tabular}{l|c|c|c}
	\hline
	\textbf{Method}          & \textbf{Dice$_{OC}$(\%)} & \textbf{Dice$_{OD}$(\%)} & \textbf{mIoU(\%)} \\ \hline
	Base Network             &          90.11          &          95.77          &       84.41        \\
	Base Network + EGM       &          91.24          &          97.17          &       86.49        \\
	Base Network + WAM       &          91.49          &          97.07          &       86.62        \\ \hline
	Base Network + EGM + WAM &     \textbf{93.14}      &     \textbf{97.52}      &   \textbf{87.92}   \\ \hline
\end{tabular}}
\label{tab:ablation}
\end{center}
\end{table}

\subsection{Ablation Study}

To evaluate the contributions of each component of the proposed method, we conduct experiments with different settings on the Drishti-GS dataset. As shown in Table~\ref{tab:ablation}, we choose the encoder-decoder network shown in Fig.~\ref{fig_network} as the base network, which achieves 90.11\%, 95.77\% and 84.41\% in terms of Dice$_{OC}$, Dice$_{OD}$ and mIoU, respectively. When we append the proposed EGM, it yields results of 91.24\%/97.17\%/86.49\% (Dice$_{OC}$/Dice$_{OD}$/mIoU). This dramatically outperforms the base network, with only a small addition to the computational cost, proving that edge information is of vital importance for segmentation.
To study the effect of the WAM, we append the WAM to the base network, without concatenating the edge features to base network. With the same training settings, this approach achieves performances of 91.49\%/97.07\%/86.62\%, compared to the base network. The obvious performance gains for all three metrics illustrate the efficiency of the proposed the WAM. Finally, our whole ET-Net, with both EGM and WAM, obtains the best performance on the Drishti-GS \textit{test} set.

\section{Conclusion}
\label{sec:conclusion}

In this paper, we propose a novel Edge-aTtention Guidance network (ET-Net) for general medical image segmentation. By assuming that edge detection and region segmentation are mutually beneficial, we have proposed the Edge Guidance Module to detect object edges and generate edge-attention representations that contain sufficient edge information. Moreover, a Weighted Aggregation Module has been employed to highlight the valuable features of high-level layers, which are combined with the edge representations, to guide the final segmentation. Experiments on various medical imaging tasks have demonstrated the superiority of our proposed ET-Net compared to other state-of-the-art methods. In future work, we will extend our approach to 3D segmentation on CT and MRI volumes.

\bibliographystyle{splncs04}
\bibliography{paper}

\begin{thebibliography}{10}
\providecommand{\url}[1]{\texttt{#1}}
\providecommand{\urlprefix}{URL }
\providecommand{\doi}[1]{https://doi.org/#1}

\bibitem{Aquino2010}
Aquino, A., Gegundez-Arias, M.E., Marin, D.: {Detecting the optic disc boundary
  in digital fundus images using morphological, edge detection, and feature
  extraction techniques}. IEEE TMI  (2010)

\bibitem{Berman_2018_CVPR}
Berman, M., Rannen~Triki, A., Blaschko, M.B.: The lov\'{a}sz-softmax loss: A
  tractable surrogate for the optimization of the intersection-over-union
  measure in neural networks. In: CVPR (2018)

\bibitem{multitask}
Chen, H., Qi, X., et~al.: {DCAN:} deep contour-aware networks for accurate
  gland segmentation. In: CVPR (2016)

\bibitem{Cheng2013}
Cheng, J., Liu, J., et~al.: {Superpixel classification based optic disc and
  optic cup segmentation for glaucoma screening}. IEEE TMI  (2013)

\bibitem{mnet2018fu}
Fu, H., Cheng, J., et~al.: {Joint Optic Disc and Cup Segmentation Based on
  Multi-Label Deep Network and Polar Transformation}. IEEE TMI  (2018)

\bibitem{deepvessel2016}
Fu, H., Xu, Y., et~al.: {DeepVessel: Retinal Vessel Segmentation via Deep
  Learning and Conditional Random Field}. In: MICCAI (2016)

\bibitem{Gu2019}
Gu, Z., Cheng, J., et~al.: {CE-Net: Context Encoder Network for 2D Medical
  Image Segmentation}. IEEE TMI  (2019)

\bibitem{resnet2016he}
He, K., Zhang, X., et~al.: Deep residual learning for image recognition. In:
  CVPR (2016)

\bibitem{mc}
Jaeger, S., Candemir, S., et~al.: Two public chest x-ray datasets for
  computer-aided screening of pulmonary diseases. QIMS  (2014)

\bibitem{lung2015}
Mansoor, A., Bagci, U., et~al.: {Segmentation and Image Analysis of Abnormal
  Lungs at CT: Current Approaches, Challenges, and Future Trends}.
  Radiographics  (2015)

\bibitem{MOCCIA201871}
Moccia, S., Momi, E.D., et~al.: Blood vessel segmentation algorithms — review
  of methods, datasets and evaluation metrics. CMPB  (2018)

\bibitem{Ronneberger2015}
Ronneberger, O., Fischer, P., Brox, T.: {U-Net: Convolutional Networks for
  Biomedical Image Segmentation}. In: MICCAI (2015)

\bibitem{fcn}
Shelhamer, E., Long, J., Darrell, T.: Fully convolutional networks for semantic
  segmentation. TPAMI  (2017)

\bibitem{Drishti}
Sivaswamy, J., Krishnadas, S.R., et~al.: Drishti-gs: Retinal image dataset for
  optic nerve head(onh) segmentation. In: IEEE ISBI (2014)

\bibitem{DRIVE}
Staal, J., Abr{\`{a}}moff, M.D., et~al.: Ridge-based vessel segmentation in
  color images of the retina. IEEE TMI  (2004)

\bibitem{levelset2003}
{Tsai}, A., {Yezzi}, A., et~al.: A shape-based approach to the segmentation of
  medical imagery using level sets. IEEE TMI  (2003)

\bibitem{Wang2019tmi}
Wang, S., Yu, L., et~al.: Patch-based output space adversarial learning for
  joint optic disc and cup segmentation. IEEE TMI  (2019)

\bibitem{salient}
Wang, W., Lai, Q., et~al.: Salient object detection in the deep learning era:
  An in-depth survey. arXiv:1904.09146  (2019)

\bibitem{shen2019}
Wang, W., Shen, J., Ling, H.: A deep network solution for attention and
  aesthetics aware photo cropping. IEEE PAMI  (2019)

\end{thebibliography}
\end{document}